\pdfoutput=1

\documentclass[11pt]{article}

\usepackage{eacl2023}

\usepackage{times}
\usepackage{latexsym}
\usepackage{amsmath}
\usepackage{amssymb}
\usepackage{amsthm}
\usepackage{graphicx}
\usepackage{booktabs}
\usepackage{arydshln}
\usepackage{makecell}
\usepackage{multirow}
\usepackage{multicol}
\usepackage{enumitem}
\usepackage{bbm}
\usepackage{soul}
\usepackage[ruled, linesnumbered]{algorithm2e}
\usepackage[htt]{hyphenat}
\usepackage{framed}
\usepackage{comment}
\usepackage{placeins}
\usepackage{titlesec}

\titlespacing\section{0pt}{1.6ex plus 2pt minus 2pt}{1.0ex plus 2pt minus 2pt}
\titlespacing\subsection{0pt}{1.6ex plus 2pt minus 2pt}{1.0ex plus 2pt minus 2pt}
\titlespacing\paragraph{0pt}{0.5ex plus 2pt minus 2pt}{1em}

\theoremstyle{definition}
\newtheorem{definition}{Definition}[section]

\theoremstyle{definition}

\newtheorem{theorem}{Theorem}

\makeatletter
\def\adl@drawiv#1#2#3{%
        \hskip.5\tabcolsep
        \xleaders#3{#2.5\@tempdimb #1{1}#2.5\@tempdimb}%
                #2\z@ plus1fil minus1fil\relax
        \hskip.5\tabcolsep}
\newcommand{\cdashlinelr}[1]{%
  \noalign{\vskip\aboverulesep
           \global\let\@dashdrawstore\adl@draw
           \global\let\adl@draw\adl@drawiv}
  \cdashline{#1}
  \noalign{\global\let\adl@draw\@dashdrawstore
           \vskip\belowrulesep}}
\makeatother

\usepackage[T1]{fontenc}
\usepackage[utf8]{inputenc}
\usepackage{dblfloatfix}

\usepackage{microtype}

\newcommand{\dfmit}{DF\textsubscript{MIT}}
\newcommand{\dffrac}{DF\textsubscript{Frac}}
\newcommand{\rankdel}{Rank\textsubscript{Del}}
\newcommand{\rankins}{Rank\textsubscript{Ins}}
\newcommand{\estar}{E$^*$}

\newcommand{\ctext}[3][RGB]{%
  \begingroup
  \definecolor{hlcolor}{#1}{#2}\sethlcolor{hlcolor}%
  \hl{#3}%
  \endgroup
}

\let\vec\mathbf

\DeclareMathOperator*{\argmin}{arg\,min}

\newcommand{\s}[1]{\textsubscript{#1}}

\title{The Solvability of Interpretability Evaluation Metrics}

\author{Yilun Zhou \\ MIT CSAIL \\ \texttt{ yilun@csail.mit.edu} \And
Julie Shah\\ MIT CSAIL \\ \texttt{ julie\_a\_shah@csail.mit.edu}}

\author{Yilun Zhou \qquad Julie Shah\vspace{0.03in}\\\vspace{0.01in}MIT CSAIL\\\texttt{\{yilun,julie\_a\_shah\}@csail.mit.edu}\vspace{0.07in}\\\url{https://yilunzhou.github.io/solvability/}}

\begin{document}
\maketitle

\begin{abstract}
Feature attribution methods are popular for explaining neural network predictions, and they are often evaluated on metrics such as comprehensiveness and sufficiency. In this paper, we highlight an intriguing property of these metrics: their \textit{solvability}. Concretely, we can define the problem of optimizing an explanation for a metric, which can be solved by beam search. This observation leads to the obvious yet unaddressed question: why do we use explainers (e.g., LIME) not based on solving the target metric, if the metric value represents explanation quality? We present a series of investigations showing strong performance of this beam search explainer and discuss its broader implication: a definition-evaluation duality of interpretability concepts. We implement the explainer and release the Python \texttt{solvex} package for models of text, image and tabular domains. 
\end{abstract}

\section{Introduction}
\label{sec:intro}
For neural network models deployed in high stakes domains, the explanations for predictions are often as important as the predictions themselves. For example, a skin cancer detection model may work by detecting surgery markers \citep{winkler2019association} and an explanation that reveals this spurious correlation is highly valuable. However, evaluating the correctness (or faithfulness) of explanations is fundamentally ill-posed: because the explanations are used to help people understand the reasoning of the model, we cannot check it against the ground truth reasoning, as the latter is not available. 

\begin{figure}[!t]
    \centering
    \includegraphics[width=0.95\columnwidth]{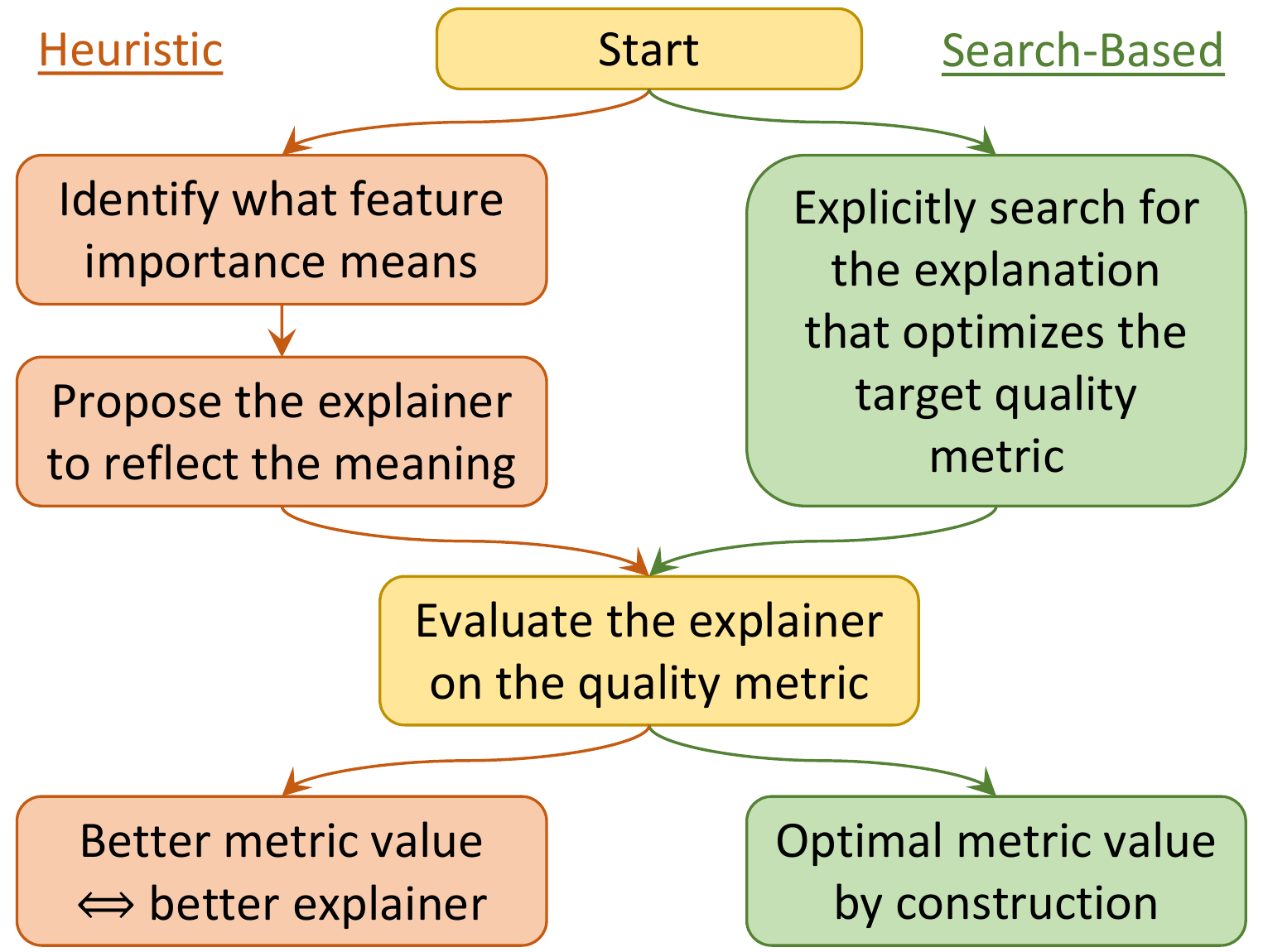}
    \caption{Left: the current process of developing new explainers. Right: the natural implication following our observation that evaluation metrics are \textit{solvable}. }
    \label{fig:fig1}
\end{figure}

As a result, correctness evaluations typically employ certain alternative metrics. For feature attribution explanations, they work under a shared principle: changing an important feature should have a large impact on the model prediction. Thus, the quality of the explanation is defined by different formulations of the model prediction change, resulting in various metrics such as comprehensiveness and sufficiency \citep{deyoung2020eraser}. To develop new explanation methods (Fig.~\ref{fig:fig1}, left), people generally identify a specific notion of feature importance (e.g., local sensitivity), propose the corresponding explainer (e.g., gradient saliency \citep{simonyan2013deep}), evaluate it on one or more metrics, and claim its superiority based on favorable results vs. baseline explainers. We call these explainers \textit{heuristic} as they are motivated by pre-defined notions of feature importance. 

In this paper, we show that all these metrics are \textit{solvable}, in that we can \textit{define} an explanation as the one that optimizes a metric value and \textit{search} for it. The obvious question is then: \textit{if we take a specific target metric to represent correctness, why don't we just search for the metric-optimal explanation (Fig.~\ref{fig:fig1}, right) but take the more convoluted route of developing heuristic explanations and then evaluating them (Fig.~\ref{fig:fig1}, left)? }

\begin{figure*}[t]
    \centering
    \includegraphics[width=\textwidth]{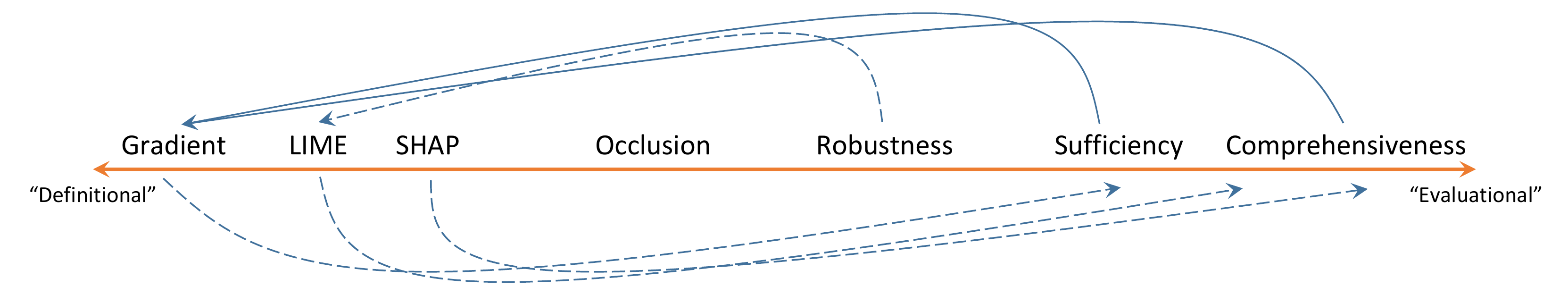}
    \vspace{-0.3in}
    \caption{A definition-evaluation spectrum for various interpretability concepts currently as perceived by the community (see App.~\ref{app:spectrum} for some justification). The proposed solvability property can move evaluational concepts towards the definitional side, for which we explore two in the paper (solid arrows). The more general definition-evaluation duality opens up new opportunities to move other concepts around (dashed arrows). }
    \vspace{-0.2in}
    \label{fig:spectrum}
\end{figure*}

There are several possible reasons. First, the optimization problem may be so hard that we cannot find an explanation better than the heuristic ones. The bigger concern, however, is that of Goodhart's Law. In other words, as soon as a metric is used in explicit optimization, it ceases to be a good metric. Concretely, the explanation may overfit to the particular metric and perform much worse on closely related ones \citep{chan2022comparative}, or overfit to the model and effectively adversarially attack the model when assigning word importance \citep{feng2018pathologies}. It may also perform poorly on evaluations not based on such metrics, such as ground truth alignment \citep{zhou2021feature}. 

We assess these concerns, taking the widely used comprehensiveness and sufficiency metrics \citep{deyoung2020eraser} as the optimization target. Our findings, however, largely dispel every concern. A standard beam search produces explanations that greatly outperform existing one such as LIME and SHAP on the target metric. On several other metrics, the search-based explainer also performs favorably on average. There is no strong evidence of it adversarially exploiting the model either, and it achieves competitive performances on a suite of ground truth-based evaluations. 

Thus, we advocate for wider adoptions of the explainer, which is domain-general and compatible with models on image and tabular data as well. As an engineering contribution, we release the Python \texttt{solvex} package (\textit{\textbf{solv}}ability-based \textit{\textbf{ex}}planation) and demonstrate its versatility in App.\ref{app:solvex}. 

More broadly, the solvability phenomenon is one facet of the definition-evaluation duality, which asserts an equivalence between definitions and evaluations. Solvability recognizes that for each evaluation metric, we can define explainer that performs optimally on this metric. Conversely, for each explainer, we can also come up with an evaluation metric that ranks this explainer on top -- a straightforward one would be the negative distance between the explanation under evaluation and the ``reference explanation'' generated by the explainer. 

While the community has mostly agreed on a spectrum on which various interpretability concepts (Fig.~\ref{fig:spectrum}) are located, duality allows every concept to be moved freely on the scale. We explored two particular movements as represented by the solid arrows, but the more general investigation of this operation could be of both theoretical and practical interest. In addition, given that definitions and evaluations are really two sides of the same coin, we need to reflect how to best evaluate explanations. Sec.~\ref{sec:discussion} argues to measure their \textit{demonstrable utilities} in downstream tasks, and present potential ways and ideas to better align the interpretability research with such goals.

\section{Background and Related Work}
\label{sec:background}

In this section, we give a concise but unified introduction to the popular feature attribution explainers and evaluation metrics studied in this paper. 

\subsection{Feature Attribution Explainers}

We focus on feature attribution explanations, which explains an input $x=(x_1, ..., x_L)$ by a vector $e=(e_1, ..., e_L)$ where $e_l$ represents the ``contribution'' of $x_l$ to the prediction. Many different definitions for contribution have been proposed and we consider the following five. 

\begin{itemize}[leftmargin=*, itemsep=0pt, parsep=0.3ex, topsep=0.1ex]
    \item \textbf{Vanilla gradient (Grad)} \citep{simonyan2013deep, li2015visualizing} is the L2 norm of gradient of the prediction (in logit, following standard practice) with respect to the token embedding. 
    \item \textbf{Integrated gradient (IntG)} \citep{sundararajan2017axiomatic} is the path integral of the embedding gradient along the line segment from the zero embedding value to the actual value. 
    \item \textbf{LIME} \citep{ribeiro2016should} is the coefficient of a linear regression in the local neighborhood. 
    \item \textbf{SHAP} \citep{lundberg2017unified} computes the Shapley value \citep{roth1988shapley} for each word. 
    \item \textbf{Occlusion (Occl)} \citep{li2016understanding} is the change in prediction when a word is removed from the input while all other words remain. 
\end{itemize}

\subsection{Feature Attribution Evaluations}

Naturally, different definitions result in different explanation values. As findings \citep[e.g.,][]{adebayo2018sanity, nie2018theoretical} suggest that some explanations are not correct (i.e., faithfully reflecting the model's reasoning process), many evaluations are proposed to quantify the correctness of different explanations. Not having access to the ground truth model working mechanism (which is what explanations seek to reveal in the first place), they are instead guided by one principle: changing an important feature (as judged by the explanation) should have a large impact on the prediction, and the magnitude of the impact is taken as explanation quality. However, there are different ways to quantify the impact, leading to different evaluations, and we consider six in this paper. 

Let $f: \mathcal X\rightarrow \mathbb R$ be a function that we want to explain, such as the probability of the target class. For an input $x=(x_1, ..., x_L)$ of $L$ words, according to an explanation $e=(e_1, ..., e_L)$, we can create a sequence of $L+1$ input deletions $\tilde x^{(0)}_e, \tilde x^{(1)}_e, ..., \tilde x^{(L)}_e$ where $\tilde x^{(l)}_e$ is the the input but with $l$ most important features removed. Thus, we have $\tilde x^{(0)}_e=x$ and $\tilde x^{(L)}_e$ being the empty string.\footnote{We define feature removal as the literal deletion of the word from the sentence, which is a popular practice. Other methods replace the token with [UNK], [MASK] or zero embedding, are more sophisticated such as performing BERT mask filling \citep{kim2020interpretation}. While our current approach could lead to out-of-distribution instances, we adopt it due to its popularity. A thorough investigation for the best strategy is orthogonal to our paper and beyond its scope. } The \textbf{comprehensiveness} $\kappa$ \citep{deyoung2020eraser} is defined as 
\begin{align}
    \kappa(x, e) = \frac{1}{L+1} \sum_{l=0}^L f(x) - f(\tilde x^{(l)}_e). \label{eq:comp}
\end{align}
It measures the deviation from the original model prediction when important features (according to $e$) are successively removed, and therefore a larger value is desirable. It was also proposed for computer vision models as the area over perturbation curve (AoPC) by \citet{samek2016evaluating}. 

Analogously, we can define the sequence of input insertions $\widehat x^{(0)}_e, \widehat x^{(1)}_e, ..., \widehat x^{(L)}_e$, where $\widehat x^{(l)}_e$ is the input with the $l$ most important features present. Thus, $\widehat x^{(0)}_e$ is the empty string and $\widehat x^{(L)}_e=x$, but otherwise the sequences of input insertions and deletions do not mirror each other. The \textbf{sufficiency} $\sigma$ \citep{deyoung2020eraser} is defined as 
\begin{align}
    \sigma(x, e) = \frac{1}{L+1} \sum_{l=0}^L f(x) - f(\widehat x^{(l)}_e). \label{eq:suff}
\end{align}
It measures the gap to the original model prediction that remains (i.e., convergence to the model prediction) when features are successively inserted from the most important to the least. Therefore, a smaller value is desirable. 

Another interpretation of prediction change just considers decision flips. Let $g: \mathcal X\rightarrow \{0, ..., K\}$ be the function that outputs the most likely class of an input. The \textbf{decision flip by removing the most important token} \citep{chrysostomou2021improving} is defined as 
\begin{align}
    \mathrm{DF_{MIT}}(x, e) = \mathbbm 1_{g(\tilde x^{(1)}_e)\neq g(x)}, 
\end{align}
which measures whether removing the most important token changes the decision. Across a dataset, its average value gives the overall decision flip rate, and a higher value is desirable. 

The \textbf{fraction of token removals for decision flip} \citep{serrano2019attention} is defined as
\begin{align}
    \mathrm{DF_{Frac}}(x, e) = \frac{\argmin_{l} g(\tilde x^{(l)}_e)\neq g(x)}{L}, 
\end{align}
and we define $\mathrm{DF_{Frac}}=1$ if no value of $l$ leads to the decision flip. This metric represents the fraction of feature removals that is needed to flip the decision, and hence a lower value is desirable. 

Last, two metrics evaluate correlations between model prediction and feature importance. For $x$ and $e$, we define the sequence of marginal feature deletions $x_-^{(1)}, ..., x_{-, e}^{(L)}$ such that $x{_, e}-^{(l)}$ is original input with only the $l$-th important feature removed. The \textbf{deletion rank correlation} \citep{alvarez2018towards} is defined as
\begin{gather}
    \delta_f = [f(x)-f(x_{-, e}^{(1)}), ..., f(x)-f(x_{-, e}^{(L)})], \\
    \mathrm{Rank_{Del}}(x, e) = \rho(\delta_f, e), 
\end{gather}
where $\rho(\cdot, \cdot)$ is the Spearman rank correlation coefficient between the two input vectors. Intuitively, this metric asserts that suppressing a more important feature should have a larger impact to the model prediction. A higher correlation is desirable. 

The \textbf{insertion rank correlation} \citep{luss2021leveraging} is defined as
\begin{gather}
v = [f(\tilde x^{(L)}), ..., f(\tilde x^{(0)})], \\
\mathrm{Rank_{Ins}}(x, e) = \rho(v, [0, ..., L]), 
\end{gather}
and recall that $\tilde x^{(L)}_e, ..., \tilde x^{(0)}_e$ is the sequence of inputs with increasingly more important features inserted, starting from the empty string $\tilde x^{(L)}$ to the full input $\tilde x^{(0)}$. This metric asserts that the model prediction on this sequence should increase monotonically to the original prediction. Also a higher correlation is desirable.

Related to our proposed notion of solvability is the phenomenon that some metric values seem to favor some explainers \citep{pham2021double, ju2022logic}. While it is often used to argue \textit{against} the use of certain evaluations, we take this idea to the extreme, which culminates in the solvability property, and find that metric-solving (Def.~\ref{def:solve}) explanations from some metrics can be high-quality. 

\section{The Solvability of Evaluation Metrics}
Now we establish the central observation of this paper: the solvability of these evaluation metrics. Observe that each evaluation metric, e.g., comprehensiveness $\kappa$, is defined on the input $x$ and the explanation $e$, and its computation only uses the model prediction function $f$ (or $g$ derived from $f$ for the two decision flip metrics). In addition, the form of feature attribution explanation constrains $e$ to be a vector of the same length as $x$, or $e\in \mathbb R^L$. 

Without loss of generality, we assume that the metrics are defined such that a higher value means a better explanation (e.g., redefining the sufficiency to be the negative of its original form). We formalize the concept of solvability as follows: 
\begin{definition}
For a metric $m$ and an input $x$, an explanation $e^*$ \textit{solves} the metric $m$ if $m(x, e^*)\geq m(x, e)$ for all $e\in \mathbb R^L$. We also call $e^*$ the $m$-\textit{solving} explanation. \label{def:solve}
\end{definition}
Notably, there are already two explanation-solving-metric cases among the ones in Sec.~\ref{sec:background}. 

\begin{theorem}
The occlusion explainer solves the \dfmit{} and \rankdel{} metrics. \label{thm:occl}
\end{theorem}

The proof follows from the definition of the explainer and the two metrics. Occlusion explainer defines token importance as the prediction change when each the token is individually removed, thus the most important token is the one that induces the largest change, which makes it most likely to flip the decision under \dfmit{}. In addition, because token importance is defined as the model prediction change, its rank correlation with the latter (i.e., \rankdel{}) is maximal at 1.0. 

Thm.~\ref{thm:occl} highlights an important question: if we take \dfmit{} or \rankdel{} as the metric (i.e., indicator) of explanation quality, why should we consider any other explanation, when the occlusion explanation provably achieves the optimum? A possible answer is that the metrics themselves are problematic. For example, one can argue that the \dfmit{} is too restrictive for overdetermined input: when redundant features (e.g., synonyms) are present, removing any individual one cannot change the prediction, such as for the sentiment classification input of ``This movie is great, superb and beautiful.'' 

Nonetheless, the perceived quality of a metric can be loosely inferred from its adoption by the community, and the comprehensiveness and sufficiency metrics \citep{deyoung2020eraser} are by far the most widely used. They overcome the issue of \dfmit{} by also considering inputs with more than one token removed. Since a metric is scalar-valued, we combine comprehensiveness $\kappa$ and sufficiency $\sigma$ into comp-suff difference $\Delta$, defined as (recall that a \textit{lower} sufficiency value is better): 
\begin{align}
    \Delta(x, e) = \kappa(x, e) - \sigma(x, e). 
\end{align}

Again, we face the same question: if $\Delta$ is solvable, why should \textit{any} heuristic explainers be used instead of the $\Delta$-solving $e^*$? In the next two sections, we seek to answer it by first proposing a beam search algorithm to (approximately) find $e^*$ and then explore its various properties. 

\section{Solving Metrics with Beam Search}
\label{sec:beam-search}
We first define two properties that are satisfied by some metrics: value agnosticity and additivity.
\begin{definition}
\label{def:value-agnostic}
For an input $x=(x_1, ..., x_L)$ with explanation $e=(e_1, ..., e_L)$, we define the ranked importance as $r(x_l) = |\{e_i: e_i\leq e_l, 1\leq i \leq L\}|$. In other word, the $x_l$ with $r(x_l)=L$ is the most important, and that with $r(x_l)=1$ is the least. A metric $m$ is \textit{value-agnostic} if for all $e_1$ and $e_2$ that induce the same ranked importance, we have 
\begin{align}
    m(x, e_1) = m(x, e_2). 
\end{align}
\end{definition}
A value-agnostic metric has at most $L!$ unique values across all possible explanations for an input of length $L$. Thus, in theory, an exhaustive search over the $L!$ permutations of the list $[1, 2, ..., L]$ is guaranteed to find the $e^*$ that solves the metric. 

\begin{definition}
A metric $m$ is \textit{additive} if it can be written in the form of 
\begin{align}
    m(x, e) = \sum_{l=0}^{L} h(x, e^{(l)}), 
\end{align}
for some function $h$, where $e^{(l)}$ reveals the attribution values of $l$ most important features according to $e$ but keeps the rest inaccessible. 
\end{definition}

\begin{theorem}
Comprehensiveness, sufficiency and their difference are value-agnostic and additive. 
\end{theorem}
The proof is straightforward, by observing that both $\tilde x^{(l)}$ and $\widehat x^{(l)}$ can be created from $x$ and the ordering of $e^{(l)}$. In fact, all metrics in Sec.~\ref{sec:background} are value-agnostic (but only some are additive). 

A metric satisfying these two properties admits an efficient beam search algorithm to approximately solve it. As $e^{(l)}$ can be considered as a partial explanation that only specifies the top-$l$ important features, we start with $e^{(0)}$, and try each feature as most important obtain $e^{(1)}$. With beam size $B$, if there are more than $B$ features, we keep the top-$B$ according to the partial sum. This extension procedure continues until all features are added, and top extension is then $e^*$. Alg.~\ref{alg:beam-search} documents the procedure, where $\mathrm{ext}(e, v)$ extends $e$ and returns a set of explanations, in which each new one has value $v$ on one previously empty entry of $e$. Finally, note that $e^*$ generated on Line 8 has entry values in $\{1, ..., L\}$, but some features may contribute \textit{against} the prediction (e.g., ``This movie is truly innovative although slightly \ul{cursory}.''). Thus, we post-process $e^*$ by shifting all values by $k$ such that the new values (in $\{1-k, L-k\}$) maximally satisfy the sign of marginal contribution of each word (i.e., the sign of the occlusion saliency). 

\begin{algorithm}[!htb]
\textbf{Input}: beam size $B$, metric $m$, sentence $x$ of length $L$\; 
Let $e^{(0)}$ be an empty length-$L$ explanation\;
\texttt{beams} $\gets \{e^{(0)}\}$\;
\For{$l = 1, ..., L$}{
\texttt{beams} $\displaystyle \gets \bigcup_{e \, \in \, \texttt{beams}} \mathrm{ext}(e, L-l+1)$\;
\texttt{beams} $\gets \mathrm{choose\_best}(\texttt{beams}, B)$\;
}
$e \gets \mathrm{choose\_best}(\texttt{beams}, 1)$\;
$e^* \gets \mathrm{shift}(e)$\;
\textbf{return} $e^*$\; 
\caption{Beam search for finding $e^*$.}
\label{alg:beam-search}
\end{algorithm}

Without the additive property, beam search is not feasible due to the lack of partial metric values. However, \citet{zhou2021towards} presented a simulated annealing algorithm \citep{kirkpatrick1983optimization} to search for the optimal data acquisition order in active learning, and we can use a similar procedure to search for the optimal feature importance order. If the metric is value-sensitive, assuming differentiability with respect to the explanation value, methods such as gradient descent can be used. Since we focus on comprehensiveness and sufficiency in this paper, the development and evaluation of these approaches are left to future work. 

\section{Experiments}
We investigate various properties of the beam search explainer vs. existing heuristic explainers, using the publicly available \href{https://huggingface.co/textattack/roberta-base-SST-2}{textattack/roberta-base-SST-2} model on the SST dataset \citep{socher2013recursive} as a case study. The sentiment value for each sentence is a number between 0 (very negative) and 1 (very positive), which we binarize into two classes of $[0, 0.4]$ and $[0.6, 1]$. Sentences with sentiment values in middle are discarded. The average sentence length is 19, making the exhaustive search impossible. We use a beam size of 100 to search for $\Delta$-solving explanation \estar{}. All reported statistics are computed on the test set. 


Fig.~\ref{fig:qual-examples} presents two explanations, with additional ones in Fig.~\ref{fig:qual-examples-app} of App.~\ref{app:extra}. While we need more quantitative analyses (carried out below) for definitive conclusions on its various properties, \estar{} explanations at least looks reasonable and is likely to help people understand the model by highlighting the high importance of sentiment-laden words. 

\vspace{-0.05in}
\begin{figure}[!htb]
\begin{framed}

\ctext[RGB]{239,206,188}{A} \ctext[RGB]{236,130,102}{worthy} \ctext[RGB]{241,142,112}{tribute} \ctext[RGB]{246,186,159}{to} \ctext[RGB]{235,211,198}{a} \ctext[RGB]{232,119,93}{great} \ctext[RGB]{245,161,130}{humanitarian} \ctext[RGB]{242,200,179}{and} \ctext[RGB]{247,178,149}{her} \ctext[RGB]{243,152,121}{vibrant} \ctext[RGB]{244,194,170}{`} \ctext[RGB]{246,170,140}{co-stars} \ctext[RGB]{224,218,215}{.} \ctext[RGB]{230,215,207}{'}\vspace{0.1in}

\ctext[RGB]{151,184,254}{So} \ctext[RGB]{135,170,252}{stupid} \ctext[RGB]{144,178,254}{,} \ctext[RGB]{200,215,239}{so} \ctext[RGB]{113,148,244}{ill-conceived} \ctext[RGB]{223,219,217}{,} \ctext[RGB]{159,190,254}{so} \ctext[RGB]{122,157,248}{badly} \ctext[RGB]{211,219,230}{drawn} \ctext[RGB]{218,220,223}{,} \ctext[RGB]{166,195,253}{it} \ctext[RGB]{187,209,247}{created} \ctext[RGB]{229,216,208}{whole} \ctext[RGB]{193,212,244}{new} \ctext[RGB]{172,200,252}{levels} \ctext[RGB]{180,205,250}{of} \ctext[RGB]{128,164,250}{ugly} \ctext[RGB]{206,217,235}{.}

\end{framed}
\vspace{-0.15in}
    \caption{Two \estar{} explanations. The shade of background color represents feature importance. }
    \vspace{-0.2in}
    \label{fig:qual-examples}
\end{figure}

\subsection{Performance on the Target Metric}
We compare \estar{} to heuristic explainers on the $\Delta$ metric, with results shown in Tab.~\ref{tab:comp-suff-table} along with the associated $\kappa$ and $\sigma$. A random explanation baseline is included for reference. We can see that \estar{} achieves the best $\Delta$, often by a large margin. It also tops the ranking separately for $\kappa$ and $\sigma$, which suggests that an explanation could be optimally comprehensive and sufficient at the same time. 

\begin{table}[!htb]
    \centering
    \vspace{0.08in}
    \resizebox{0.8\columnwidth}{!}{
    \begin{tabular}{r|ccc}\toprule
        Explainer & Comp $\kappa \uparrow$ & Suff $\sigma \downarrow$ & Diff $\Delta \uparrow$ \\\midrule
        Grad & 0.327 & 0.108 & 0.218\\
        IntG & 0.525 & 0.044 & 0.481\\
        LIME & 0.682 & 0.033 & 0.649\\
        SHAP & 0.612 & 0.034 & 0.578\\
        Occl & 0.509 & 0.040 & 0.469\\\cdashlinelr{1-4}
        \estar{} & 0.740 & 0.020 & 0.720\\
        Random & 0.218 & 0.212 & 0.006\\\bottomrule
    \end{tabular}
    }
    \vspace{-0.05in}
    \caption{Comprehensiveness, sufficiency and their difference for various explainers. }
    \vspace{-0.1in}
    \label{tab:comp-suff-table}
\end{table}

\begin{figure}[!t]
    \centering
    \includegraphics[width=\columnwidth]{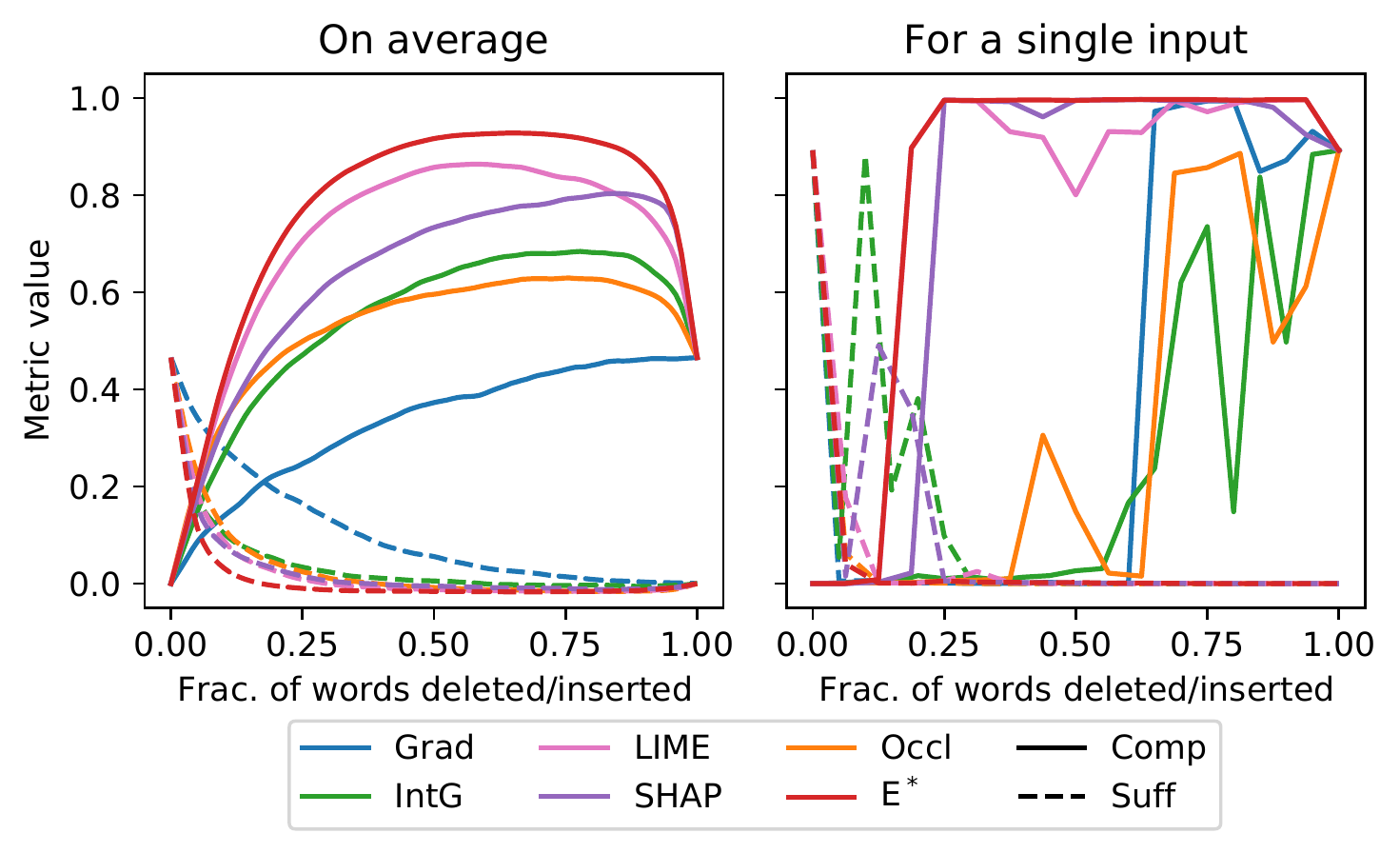}
    \vspace{-0.3in}
    \caption{Comprehensiveness and sufficiency curves for the beam search optimal explainer vs. others. }
    \vspace{-0.2in}
    \label{fig:comp-suff}
\end{figure}

To visually understand how the model prediction changes during feature removal and insertion, we plot in Fig.~\ref{fig:comp-suff} the values of $f(x) - f(\tilde x^{(l)}_e)$ and $f(x) - f(\widehat x^{(l)}_e)$ (i.e., the summands in Eq.~\ref{eq:comp} and \ref{eq:suff}), as a function of $l/L$. The left panel shows the curves averaged across all test set instances, and the right panel shows those for a specific instance. $\kappa$ and $\sigma$ are thus the areas under the solid and dashed curves respectively. The curves for \estar{} dominate the rest, and, on individual inputs, are also much smoother than those for other explanations. 

One concern for beam search is its efficiency, especially compared to those that only require a single pass of the model such as the vanilla gradient. However, we note that explanations, unlike model predictions, are rarely used in real-time decision making. Instead, they are mostly used for debugging and auditing purposes, and incurring a longer generation time to obtain a higher-quality explanation is often beneficial. On a single RTX3080 GPU card without any in-depth code optimization, the metric values and time costs for various beam sizes are presented in Tab.~\ref{tab:beam-size-ablation}, with statistics for the best explainer LIME also listed for comparison. 

\begin{table}[!b]
    \centering
    \vspace{-0.1in}
    \resizebox{\columnwidth}{!}{
    \begin{tabular}{c|llllll|l}\toprule
        $B$ & 1 & 5 & 10 & 20 & 50 & 100 & LIME \\\midrule
        $\kappa$ & 0.717 & 0.731 & 0.734 & 0.736 & 0.739 & 0.740 & 0.682 \\
        $\sigma$ & 0.020 & 0.020 & 0.020 & 0.020 & 0.020 & 0.020 & 0.033 \\
        $\Delta$ & 0.697 & 0.711 & 0.714 & 0.716 & 0.719 & 0.720 & 0.649 \\
        $T$ & 0.38 & 0.77 & 1.15 & 1.72 & 2.85 & 4.37 & 4.75 \\\bottomrule
    \end{tabular}
    }
    \caption{Effect of beam size $B$ on $\kappa, \sigma, \Delta$ and computation time $T$ (in seconds), compared against the statistics of the best heuristic explainer LIME. }
    \label{tab:beam-size-ablation}
\end{table}

Expectedly, the metric values increase with increasing beam size, but the improvement is meager after 10 beams. More importantly, beam search is not slow -- it is still faster than LIME even with 100 beams, and the single-beam version outperforms LIME by a decent margin while being more than 10 times faster. Thus, these results establish that \textit{if we take comprehensiveness and sufficiency as the quality metrics}, there is really no reason not to use the beam search explainer directly.

\subsection{Performance on Other Metrics}

Sec.~\ref{sec:background} lists many metrics that all operationalize the same principle that changing important features should have large impact on model prediction, but in different ways. A potential argument against the explicit beam search optimization is the fulfillment of Goodhart's Law: \estar{} overfits to the metric by exploiting its realization (i.e., Eq.~\ref{eq:comp} and \ref{eq:suff}) of this principle and not truly reflecting its ``spirit.''

To establish the legitimacy of this opposition, we evaluate all the explainers on the remaining four metrics in Sec.~\ref{sec:background}, and present the results in Tab.~\ref{tab:other-metrics}. 

\begin{table}[!htb]
    \centering
    \resizebox{\columnwidth}{!}{
    \begin{tabular}{r|ccccc}\toprule
        Explainer & DF\s{MIT}$\uparrow$ & DF\s{Frac}$\downarrow$ & Rank\s{Del}$\uparrow$ & Rank\s{Ins}$\uparrow$ \\\midrule
        Grad & 10.5\% & 54.5\% & 0.162 & 0.521 \\
        IntG   & 16.9\% & 39.6\% & 0.369 & 0.468 \\
        LIME & 25.5\% & 28.1\% & 0.527 & 0.342 \\
        SHAP & 23.0\% & 36.1\% & 0.369 & 0.458 \\
        Occl & 26.4\% & 40.6\% & 1.000 & 0.396 \\\cdashlinelr{1-5}
        \estar{}  & 25.0\% & 25.2\% & 0.438 & 0.423 \\
        Random & \phantom{2}3.4\% & 72.3\% & 0.004 & 0.599 \\\bottomrule
    \end{tabular}
    }
    \caption{Performance on non-target metrics of the beam search optimal explainer vs. others.}
    \label{tab:other-metrics}
\end{table}

Since the occlusion explainer solves \dfmit{} and \rankdel{} (Thm.~\ref{thm:occl}), it ranks the best on these two metrics, as expected. Nonetheless, \estar{} still ranks competitively on these two metrics and comes out ahead on \dffrac{}. The only exception is \rankins{}, on which the random explanation surprisingly performs the best. We carefully analyze it in App.~\ref{app:rank-ins} and identify a fundamental flaw in this metric. 

Last, note that we can also incorporate any of these metrics into the objective function (which already contains two metrics: $\kappa$ and $\sigma$), and search for \estar{} that performs overall the best, if so desired. We leave this investigation to future work.

\subsection{Explainer ``Attacking'' the Model}
Another concern is that the search procedure may overfit to the model. Specifically, removing a word $w$ in a partial sentence $\tilde x^{(l)}_e$ drastically changes the model prediction but does not have the same effect for most other $\tilde x^{(l')}_e$. This makes \estar{} assign $w$ an overly high attribution, as $w$ only happens to have a high impact in one particular case. By contrast, explainers like LIME and SHAP automatically avoid this issue by computing the average contribution of $w$ on many different partial sentences. 

We test this concern by locally perturbing the explanation. If \estar{} uses many such ``adversarial attacks,'' we should expect its metric values to degrade sharply under perturbation, as the high-importance words (according to \estar{}) will no longer be influential in different partial sentence contexts. 

To perturb the explanation, we first convert each explanation $e$ to its ranked importance version $e_r$ using $r(\cdot)$ in Def.~\ref{def:value-agnostic}, which does not affect any metric as they are value-agnostic. Then we define the perturbed rank by adding to each entry of $e_r$ an independent Gaussian noise: $e_r' = e_r + n$ with $n\sim \mathcal N(\vec 0, s^2)$. Thus, two words $x_i$ and $x_j$ with $r(x_i) > r(x_j)$ have their ordering switched if $r(x_i)\allowbreak - r(x_j) < n(x_j)-n(x_i)$. A visualization of the switching with different $s$ is in Fig.~\ref{fig:crossing} of App.~\ref{app:crossing}.

\begin{figure}[!t]
    \centering
    \includegraphics[width=\columnwidth]{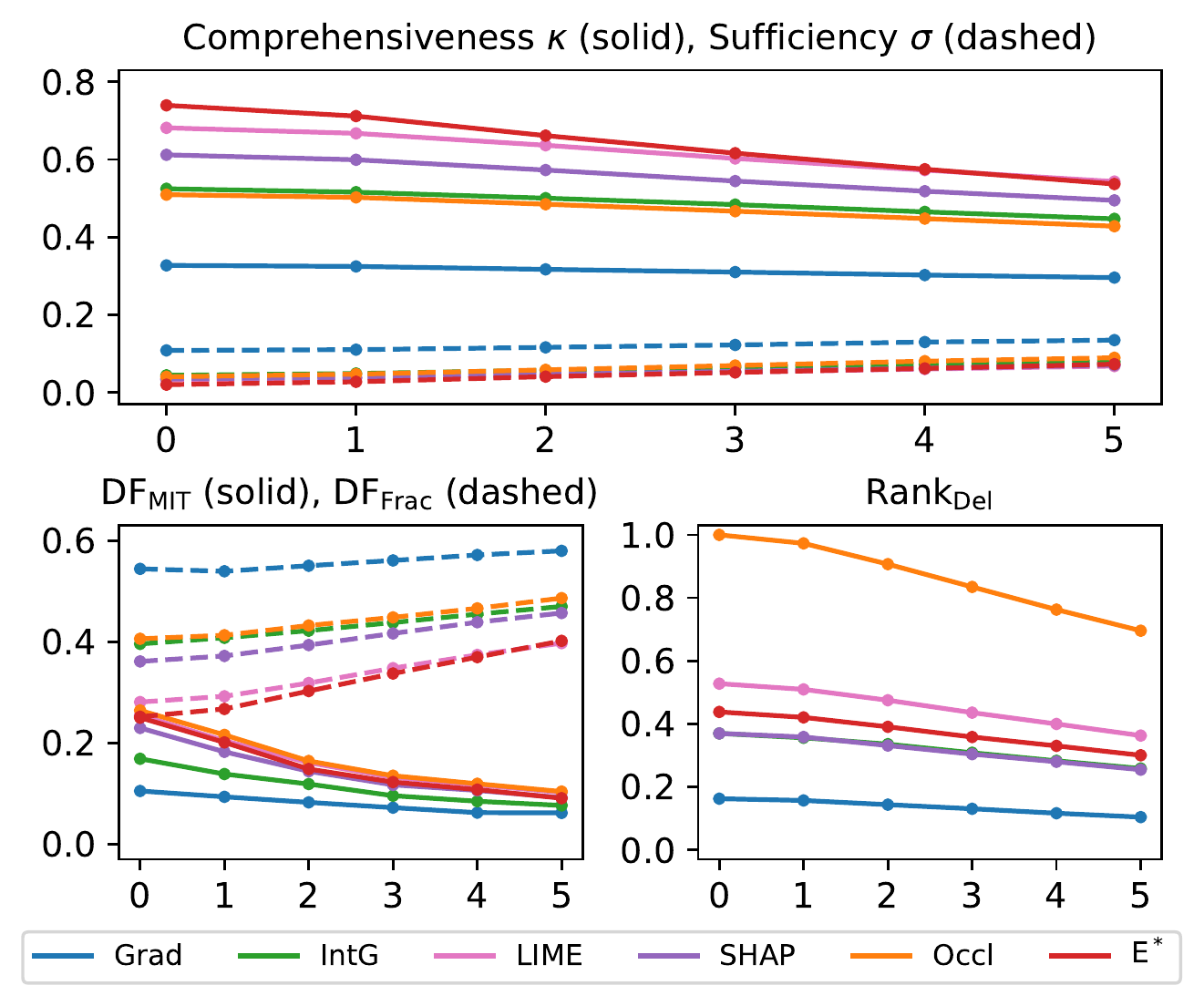}
    \vspace{-0.3in}
    \caption{Metric values for explanations under different levels of perturbation represented by $s$ on the $x$-axis. }
    \vspace{-0.2in}
    \label{fig:perturbation}
\end{figure}

Fig.~\ref{fig:perturbation} plots the metrics under different $s$ values (\rankins{} not shown due to its intrinsic issue discussed in App.~\ref{app:rank-ins}). Everything degrades to various extents. Although \estar{} degrades slightly faster than the rest on $\kappa$ and \dffrac{} (and on par on others), it still achieves best results even at $s=4$, with many order switches (Fig.~\ref{fig:crossing}), and a faster degradation is reasonable anyway for metrics with better starting values (c.f. occlusion on \rankdel{}). 

The evidence suggests that there is at most a slight model overfitting phenomenon, as \estar{} remains comparable to other explainers under quite severe perturbation. Furthermore, we can incorporate perturbation robustness into metric solving to obtain an \estar{} that degrade less, similar to adversarial training \citep{madry2017towards}. We leave the exploration of this idea to future work. 

App.~\ref{app:perturbation} describes another assessment of model overfitting, though with a mild assumption and relying on word-level sentiment scores provided by the SST dataset. Similar conclusions are reached.  

\subsection{Ground Truth Recovery}
\label{sec:ground-truth}

\setcounter{table}{3}
\begin{table}[!b]
    \centering
    \resizebox{\columnwidth}{!}{
    \begin{tabular}{c|p{3.5cm}|p{3.5cm}}\toprule
        Type & $\widehat y=0$ & $\widehat y=1$\\\midrule
        \multirow{2}{*}{Short} & terrible, awful, disaster, worst, never & excellent, great, fantastic, brilliant, enjoyable \\\midrule
        Long & \makecell[l]{A total waste of time.\\Not worth the money!\\Is it even a real film? \\ Overall it looks cheap.} & \makecell[l]{I like this movie. \\ This is a great movie! \\ Such a beautiful work. \\ Surely recommend it!} \\\bottomrule
    \end{tabular}
    }
    \caption{Set of insertions for the addition type according to the new label $\widehat y$. The words are comma-separated for ``short'', and each line is one piece of text for ``long''. }
    \label{tab:addition-ground-truth}
\end{table}

\begin{table}[!tb]
    \centering
    \resizebox{0.85\columnwidth}{!}{
    \begin{tabular}{l|l|l}\toprule
        Replacement word sets & $\widehat y=0$ & $\widehat y=1$\\\midrule
        a, an, the & a & the \\
        in, on, at & in & on \\
        I, you & I & you\\
        he, she & he & she\\
        can, will, may & can & may\\
        could, would, might & could & might\\
        (all forms of \textit{be}) & is & are\\
        (all punctuation marks) & (period) & (comma)\\
        \bottomrule
    \end{tabular}
    }
    \caption{Replacement word sets and their target words. }
    \label{tab:replacement-ground-truth}
\end{table}
\setcounter{table}{6}

\setcounter{table}{5}
\begin{table*}[!t]
    \centering
    
    \resizebox{0.85\textwidth}{!}{
    \begin{tabular}{r|cccc|cccc|cccc}\toprule
        & \multicolumn{4}{c|}{Short Addition} & \multicolumn{4}{c|}{Long Addition} & \multicolumn{4}{c}{Replacement}\\
        & \multicolumn{2}{c}{Sym} & \multicolumn{2}{c|}{Asym} & \multicolumn{2}{c}{Sym} & \multicolumn{2}{c|}{Asym} & \multicolumn{2}{c}{Sym} & \multicolumn{2}{c}{Asym} \\
        Explainer & Pr$\,\uparrow$ & NR$\,\downarrow$ & Pr$\,\uparrow$ & NR$\,\downarrow$ & Pr$\,\uparrow$ & NR$\,\downarrow$ & Pr$\,\uparrow$ & NR$\,\downarrow$ & Pr$\,\uparrow$ & NR$\,\downarrow$ & Pr$\,\uparrow$ & NR$\,\downarrow$ \\\midrule
        Grad & 0.91 & 0.06 & 0.51 & 0.08 & 0.70 & 0.37 & 0.77 & 0.30 & 0.50 & 0.75 & 0.51 & 0.74\\
        IntG & 0.82 & 0.10 & 0.60 & 0.21 & 0.60 & 0.76 & 0.70 & 0.55 & 0.49 & 0.74 & 0.48 & 0.74\\
        LIME & 1.00 & 0.06 & 1.00 & 0.06 & 0.72 & 0.60 & 0.84 & 0.32 & 0.63 & 0.65 & 0.54 & 0.71\\
        SHAP & 0.98 & 0.07 & 1.00 & 0.06 & 0.61 & 0.83 & 0.75 & 0.98 & 0.65 & 0.67 & 0.62 & 0.68\\
        Occl & 1.00 & 0.06 & 1.00 & 0.06 & 0.72 & 0.59 & 0.79 & 0.42 & 0.40 & 0.80 & 0.40 & 0.85\\ \cdashlinelr{1-13}
        \estar{} & 1.00 & 0.06 & 1.00 & 0.06 & 0.67 & 0.64 & 0.92 & 0.38 & 0.60 & 0.66 & 0.54 & 0.73\\
        Random & 0.06 & 0.54 & 0.07 & 0.53 & 0.25 & 0.89 & 0.24 & 0.88 & 0.27 & 0.85 & 0.28 & 0.85\\\bottomrule
    \end{tabular}
    }
    \caption{Average values of precision and normalized rank of the ground truth correlated words for each explainer. }
    \vspace{-0.2in}
    
    \label{tab:gtdm-metrics}
\end{table*}

For a model trained on a natural dataset, its ground truth working mechanism is rarely available -- in fact, arguably the very purpose of interpretability methods is to uncover it. Thus, a series of work \citep[e.g.,][]{zhou2021feature} proposed methods to modify the dataset such that a model trained on the new dataset has to follow a certain working mechanism to achieve high performance, which allows for evaluations against the known mechanism. 

\paragraph{Ground Truth Definitions}
We construct three types of ground truths -- short additions, long additions and replacements. First, we randomize the label to $\widehat y\sim \mathrm{Unif}\{0, 1\}$ so that the original input features are \textit{not} predictive \citep{zhou2021feature}. 

For the two addition types, a word or a sentence is inserted randomly to either the beginning or the end of the input. The inserted text is randomly chosen from the the sets in Tab.~\ref{tab:addition-ground-truth}. 

For the replacement type, each word in the input is checked against the list of replacement word sets in Tab.~\ref{tab:replacement-ground-truth}, and if the word belongs to one of the set, it is changed according to the new label $\widehat y$. On average, 27\% of input words are replaced. 

We call these modifications symmetric since inputs corresponding to both $\widehat y=0$ and $\widehat y=1$ are modified. We also define the asymmetric modification, where only inputs with $\widehat y=1$ are modified, and those with $\widehat y=0$ are left unchanged. 

\paragraph{Metrics}
We use the two metrics proposed by \citet{bastings2021will}: precision and normalized rank. First, we define the ground truth correlated words. For the two addition types, they are the inserted words. In the asymmetric case, instances with $\widehat y=0$ does not have any words added, so we exclude them in metric value computation.\footnote{This also highlights an intrinsic limitation of feature attribution explanations: they cannot explain that the model predicts a class because certain features are \textit{not} present.} For the replacement type, they are the words that are in the replacement set (but not necessarily replaced). 

Let $W$ be the set of ground truth correlated words. Using ranked importance $r(\cdot)$ in Def.~\ref{def:value-agnostic}, precision and normalized rank are defined as
\begin{align*}
    \mathrm{Pr} &= |\{w\in W: r(w) > L - |W|\}| / |W|, \\
    \mathrm{NR} &= (L - \min\{r(w): w\in W\} + 1) / L. 
\end{align*}

Precision is the fraction of ground truth words among the the top-$|W|$ ranked words, and normalized rank is the lowest rank among ground truth words, normalized by the length $L$ of the input. Both values are in $[0, 1]$, and higher precision values and lower normalized rank values are better. 

\paragraph{Results} Tab.~\ref{tab:gtdm-metrics} presents the test set Pr and NR values. Many explainers including \estar{} score perfectly on short additions, but all struggle on other types. Nonetheless, \estar{} still ranks comparably or favorably to other methods. Its largest advantage shows on the asymmetric long addition, because this setup matches with the computation of $\kappa$ and $\sigma$: \estar{} finds the most important words to remove/add to maximally change/preserve the original prediction, and those words are exactly the ground truth inserted ones. For replacement and symmetric addition, the search procedure does not ``reconstruct'' inputs of the other class, and \estar{} fails to uncover the ground truth. This finding suggests a mismatch between metric computation and certain ground truth types.  

Conversely, vanilla gradient performs decently on ground truth types other than short addition, yet ranks at the bottom on most quality metrics (Tab.~\ref{tab:comp-suff-table} and \ref{tab:other-metrics}), again likely due to the mismatch. 

\begin{figure}[!b]
    \centering
    \includegraphics[width=\columnwidth]{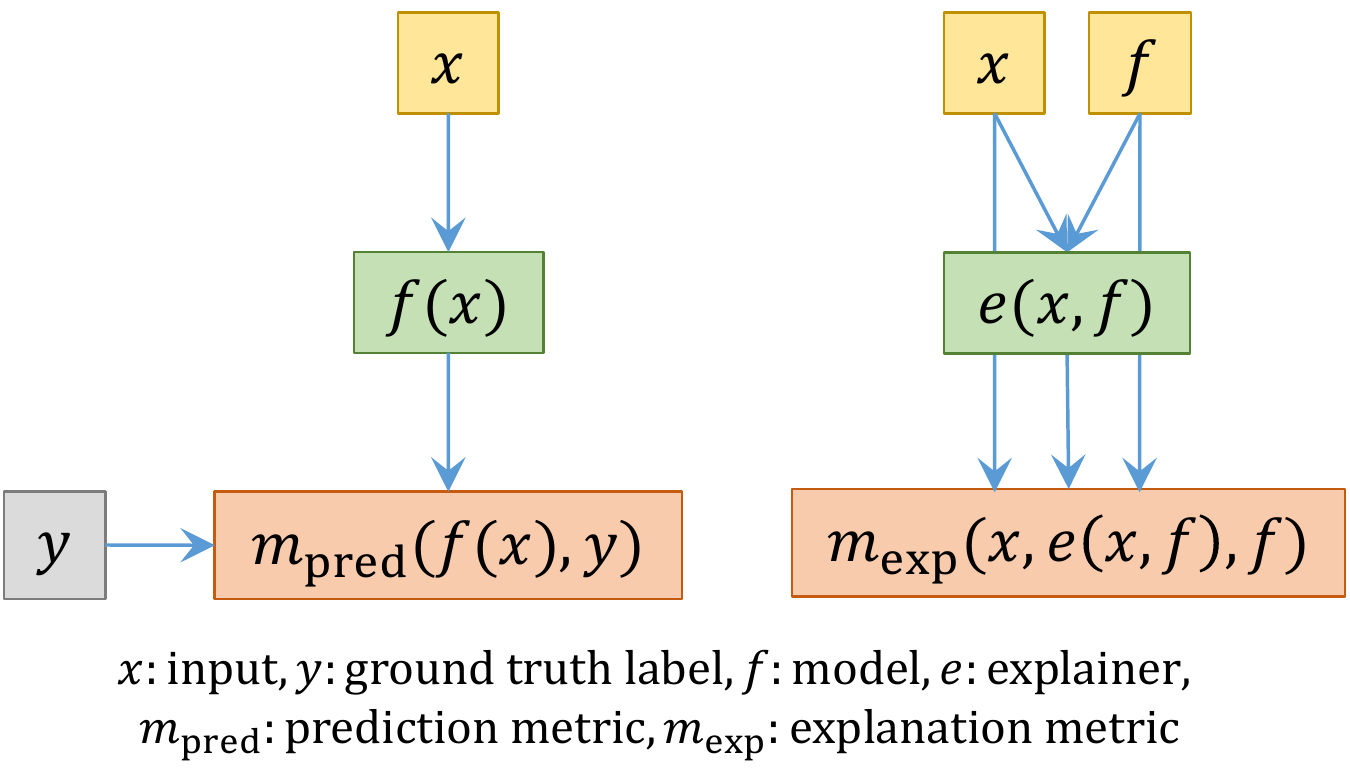}
    \caption{The complete evaluation diagrams for model predictions (left) and explanations (right). Green boxes are the model and explainer under evaluation, which have access to the information in yellow, and orange boxes are the evaluators. Notably, prediction evaluation (e.g., accuracy) uses the ground truth label $y$ not accessible to the model, but no such privileged information is used by the interpretability evaluation. }
    \label{fig:privileged-information}
\end{figure}

In fact, this evaluation is fundamentally different from the rest in its \textit{non}-solvability, specifically due to its use of privileged information. To understand this point, let us first compare the evaluation of model \textit{prediction} to that of model \textit{explanation}, as illustrated in Fig.~\ref{fig:privileged-information}. The former runs the model on the input, receives the prediction, and compares it with the ground truth label, which is emphatically \textit{not} available to the model under evaluation. By contrast, no such privileged information exists when computing interpretability metrics, allowing the explainer to directly solve them. In this ground truth recovery evaluation, we employ similar privileged information (i.e., induced ground truth model working mechanism) by dataset modification and model retraining. However, as discussed by \citet{zhou2021feature}, such evaluations are limited to the range of ground truths that could be induced.

\section{Discussion}
\label{sec:discussion}

\paragraph{Definition-Evaluation Duality} Our investigation demonstrates that some evaluation metrics can be used to find high-quality explanations, defined as the optimizers of the metrics. Conversely, we could also use any explanation definition $d$ as an evaluation metric $m$. A very simple one would be $m(x, e)\doteq -||e - d(x)||$, where $e$ is the explanation under evaluation, $d(x)$ is the ``reference explanation'' and $||\cdot||$ is a suitably chosen distance metric. It is obvious that $d(x)$ itself achieves the optimal evaluation metric value. 

Therefore, in theory, there should not be a difference of using a concept as definition vs. evaluation, but in practice, we almost always see some used mainly as definitions and others as evaluations (Fig.~\ref{fig:spectrum}). A major reason of not considering to use evaluations as definitions could be the presumed intractability of the optimization, which is experimentally refuted in this paper, as the beam search demonstrates its efficacy and efficiency. 

Conversely, why do we not see more definitions (e.g., gradients and LIME) used as evaluations? Such an attempt may sound trivial yet unjustifiable at the same time: trivial because it is equivalent to claiming that the corresponding explainer definition is the best, which is in turn a seemingly unjustifiable circular logic. 

More importantly, we motivate a new research direction opened up by the duality concept. Traditionally, definitions and evaluations have been considered and developed separately, but duality suggests that any interpretability concept can be used as both. Thus, we propose that we should focus on studying the \textit{intrinsic} properties of these concepts, independent of their usage as one or another. For example, are some concepts inherently superior for model explanations than others? How can we measure the similarity between two concepts? What does the space of these concepts look like? None of them are currently answerable due to a complete lack of formalization, but research on it could lead to a much deeper understanding of local explanations. 

\paragraph{Demonstrable Utility}
Given the duality, how should we evaluate explanations? Fundamentally, local explanations are used for model understanding \citep{zheng2021irrationality, zhou2022exsum}, and we advocate for evaluating \textit{demonstrable utility}: the presence of an explanation compared to its absence, or the newly proposed explanation compared to existing ones, should lead to a measurable difference in some practically beneficial aspect. 

For example, people use explanations to identify spurious correlation during development, audit fairness before deployment, and assist human decision makers during deployment. However, recent findings cast doubt on the feasibility of model explanations to support any of these use cases \citep{bansal2021does, jia2021role, zhou2021feature}. 

Demonstrating such utilities would bypass discussions of solvability and directly assert their usefulness~\citep{chen2022interpretable}. The examples listed here are by no means comprehensive, and a systematic taxonomy is valuable. Furthermore, it is likely that no single explainer is a one-size-fit-all solution. More refined knowledge of the strengths and weaknesses of each method in supporting different aspects of model understanding is highly desirable.

\section{Conclusion}

We study the relationship between definitions and evaluations of local explanations. We identify the \textit{solvability} property of evaluation metrics, such that for each evaluation metric, there is an explicit search procedure to find the explanation that achieves the optimal metric value. In other words, every evaluation admits a definition that \textit{solves} it. 

Compared to the current practice of defining a explainer and then evaluating it on a metric, solvability allows us to directly find the explanation that optimizes the target metric and guarantee a very favorable evaluation outcome. In this paper, we investigate the feasibility of this process. First, we propose to use beam search to find the explanation \estar{} that optimizes for comprehensiveness and sufficiency \citep{deyoung2020eraser}. Then, in a suite of evaluations, we find \estar{} performing comparably or favorably to existing explainers such as LIME. 

Therefore, for practitioners, we recommend using the proposed explainer for computing local model explanations and provide the Python \texttt{solvex} package for easy adoption (App.~\ref{app:solvex}). For researchers, we propose a definition-evaluation duality inspired by solvability, which opens up many new research directions. 

\section*{Limitations and Ethical Impact}
The focus of our paper is to investigate the search-based explanation that explicitly optimizes a target quality metric. While the results suggest that it is comparable or favorable to existing heuristic explainers on various technical aspects, its societal properties have not been studied. For example, \citet{ghorbani2019interpretation} showed that many heuristic explanations can be easily manipulated and \citet{slack2020fooling} demonstrated that discriminative models can be carefully modified such that their discrimination is hidden by heuristic explanations. It is possible that same issues exist for the search-based explanation, and thus we advise to carefully study them before deployment. 

Another limitation of this approach is that \estar{} explainer only produces rankings of feature importance, rather than numerical values of feature importance. In other words, \estar{} does not distinguish whether one feature is only slightly or significantly more important than another. By comparison, almost all heuristic explainers output numerical values (e.g., magnitude of gradient). Other than the ease of search in the ranking space than the numerical value space, we give three additional reasons. First, the utility of actual values, beyond the induced rankings, has not been well studied in the literature. In addition, many popular explanation toolkits \citep[e.g.,][]{wallace2019allennlp} even defaults to top-$k$ visualization. Last, popular evaluation metrics rarely consider values either, suggesting that there currently lack guiding principles and desiderata for these values. Moreover, if and when such value-aware metrics are widely adopted, we could augment our optimizer with them or incorporate them into a post-processing fix without affecting the ranking, similar to the shift operation done on Line 9 of Alg.~\ref{alg:beam-search}. 

\bibliographystyle{acl_natbib}
\bibliography{references}

\appendix

\onecolumn

\section{The Python \texttt{solvex} Package}
\label{app:solvex}
We release the Python \texttt{solvex} package implementing explainer in a model-agnostic manner. The project website at \url{https://yilunzhou.github.io/solvability/} contains detailed tutorials and documentation. Here, we showcase three additional use cases of the explainer. 

To explain long paragraphs, feature granularity at the level of sentences may be sufficient or even desired. \texttt{solvex} can use spaCy\footnote{\url{https://spacy.io/}} to split a paragraph into sentences and compute the sentence-level attribution explanation accordingly. As an explanation, Fig.~\ref{fig:yelp} shows an explanation for the prediction on a test instance in the Yelp dataset \citep{asghar2016yelp} made by the \href{https://huggingface.co/textattack/albert-base-v2-yelp-polarity}{albert-base-v2-yelp-polarity} model.

\begin{figure}[!htb]
    \begin{framed}
    \ctext[RGB]{242,200,179}{Contrary to other reviews, I have zero complaints about the service or the prices.} \ctext[RGB]{232,119,93}{I have been getting tire service here for the past 5 years now, and compared to my experience with places like Pep Boys, these guys are experienced and know what they're doing.} \ctext[RGB]{210,218,231}{Also, this is one place that I do not feel like I am being taken advantage of, just because of my gender.} \ctext[RGB]{189,210,246}{Other auto mechanics have been notorious for capitalizing on my ignorance of cars, and have sucked my bank account dry.} \ctext[RGB]{243,152,121}{But here, my service and road coverage has all been well explained - and let up to me to decide.} \ctext[RGB]{230,215,207}{And they just renovated the waiting room.} \ctext[RGB]{247,178,149}{It looks a lot better than it did in previous years.}

    \end{framed}
    \caption{A sentence-level explanation on a Yelp test instance. Red color indicates positive contribution. }
    \label{fig:yelp}
\end{figure}

\begin{figure}[!htb]
    \centering
    \includegraphics[width=0.9\textwidth]{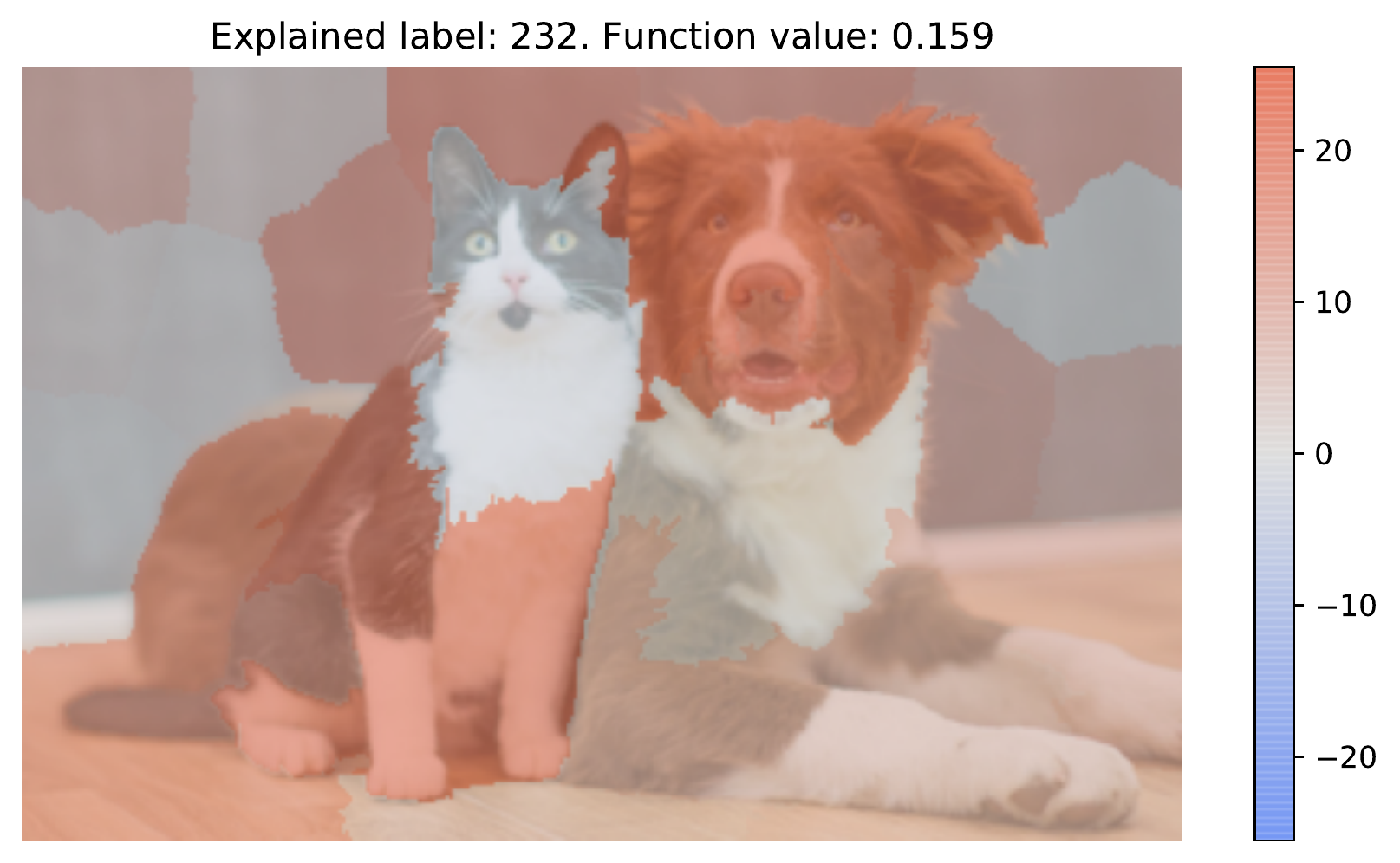}
    \caption{An explanation for the top prediction (Class 232: Border Collie, a dog breed) on an image made by a ResNet-50 model trained on ImageNet. Red color indicates positive contribution. }
    \label{fig:dog}
\end{figure}

This package can explain image predictions with superpixel features (similar to LIME \citep{ribeiro2016should}). Fig.~\ref{fig:dog} shows the explanation for the top prediction (Class 232: Border Collie, a dog breed) by the ResNet-50 \citep{he2016deep} trained on ImageNet \citep{deng2009imagenet}. 

\begin{figure}[!htb]
    \centering
    \includegraphics[width=0.7\textwidth]{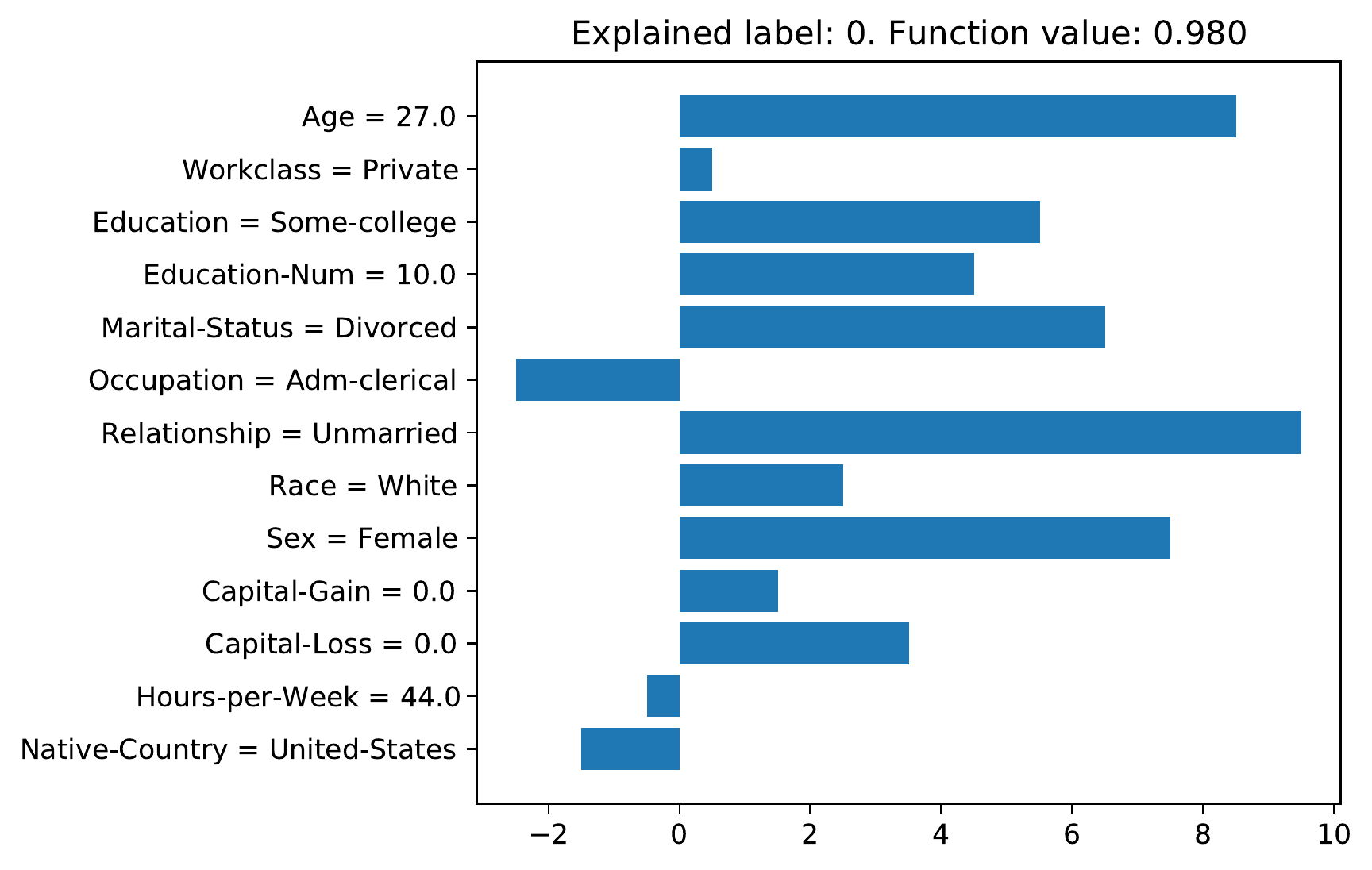}
    \caption{An explanation for the low income prediction made by a random forest model on the Adult dataset. }
    \label{fig:adult}
\end{figure}

Last, it can also explain models trained on tabular datasets with both categorical and numerical features. For a random forest model trained on the Adult dataset \citep{adult1996}, Fig.~\ref{fig:adult} shows the attribution on each feature that contributes to the class 0 (i.e., income less than or equal to $\$50K$). Note that a more positive attribution value indicates that the feature (e.g. age or relationship) contributes more to the \textit{low} income prediction. 

\section{The Definition-Evaluation Spectrum and Its Various Concepts}
\label{app:spectrum}

We describe the reasoning of assigning each concept to its location on the definition-evaluation spectrum  (Fig.~\ref{fig:spectrum}, reproduced as Fig.~\ref{fig:spectrum-2} below), as currently perceived by the community according to our understanding. Note that the discussion is unavoidably qualitative, but we hope that it illustrates the general idea of this spectrum. 

\begin{figure}[!htb]
    \centering
    \includegraphics[width=\textwidth]{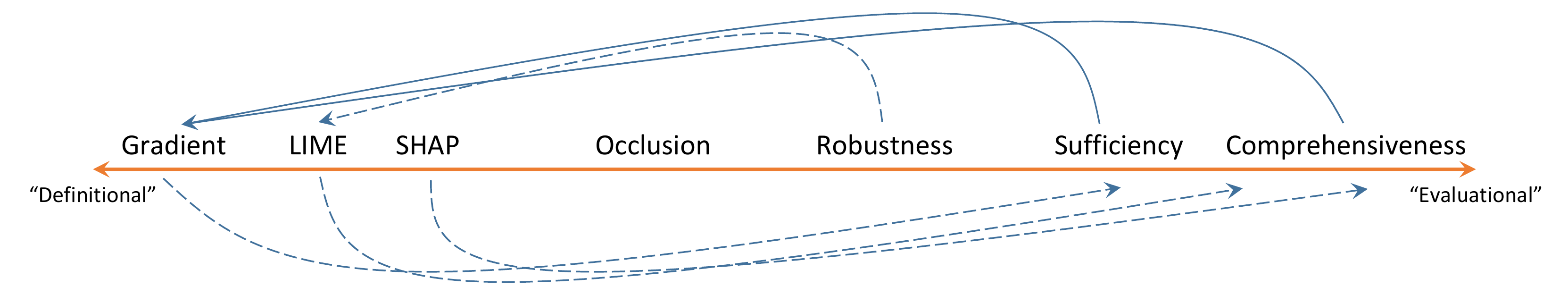}
    \caption{A definition-evaluation spectrum for various interpretability concepts, reproduced from Fig.~\ref{fig:spectrum}.}
    \label{fig:spectrum-2}
\end{figure}

We start on the definition side, where the gradient saliency \citep{simonyan2013deep, li2015visualizing} is a classic feature attribution definition but, to the best of our knowledge, has never been used in any evaluation capacity. Moving towards the evaluation side, we have LIME \citep{ribeiro2016should}, which is again used mainly to define explanations (as linear regression coefficients), but the notion of local fidelity introduced by LIME has been occasionally used to evaluate other explainers as well \citep{plumb2018model}. Similar to LIME, SHAP \citep{lundberg2017unified} defines explanations as those that (approximately) satisfy the Shapley axioms \citep{roth1988shapley}, which can also be used to evaluate how well a certain explanation performs with respect to these axioms \citep{zhang2019towards}. Next up we have the occlusion concept, which, as seen in Sec.~\ref{sec:background}, can be used as one explainer definition, Occl \citep{zeiler2014visualizing, li2016understanding}, and two (not so popular) evaluations, \dfmit{} \citep{chrysostomou2021improving} and \rankdel{} \citep{alvarez2018towards}. 

Further on the evaluation side, we now encounter concepts that are more often used for evaluations than definitions. Robustness \citep{ghorbani2019interpretation} evaluates the similarity between explanations among similar inputs and a higher degree of similarity is often more desirable \citep{alvarez2018robustness}. However, this robustness desideratum is incorporated explicitly into some explainers, such as via the noise aggregation in SmoothGrad \citep{smilkov2017smoothgrad}. On the right-most end we have sufficiency and comprehensiveness \citep{deyoung2020eraser}, which evaluates whether keeping a small subset of features could lead to the original model prediction, or removing it could lead to a large drop in model prediction. They are arguably the most popular among various evaluation metrics, and have been repeatedly proposed under different names such as the area over perturbation curve (AoPC) \citep{samek2016evaluating} and insertion/deletion metrics \citep{petsiuk2018rise}. Using such these two ideas for definitions are relatively rare, with one notable exception of smallest sufficient/destroying regions (SSR/SDR) proposed by \citep{dabkowski2017real}. 

Overall, it is clear that the community considers certain concepts more for definitions and others more for evaluations, which motivates the investigation for this paper and future work: can we swap the definition/evaluation roles, and if so, what are the implications? 

\section{Additional Qualitative Examples of the \estar{} Explanation}
\label{app:extra}

\begin{figure}[!htb]
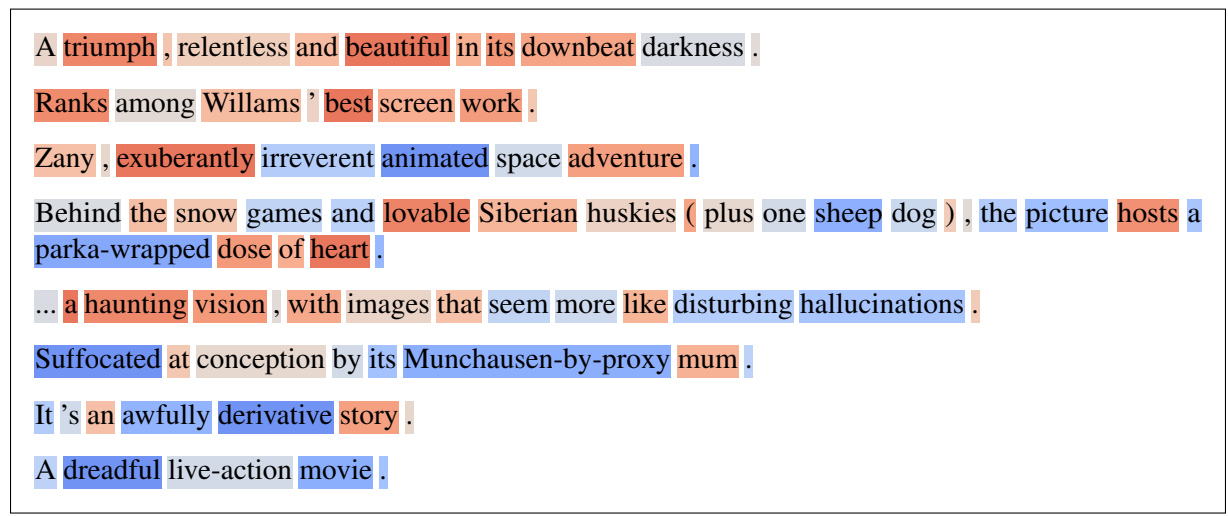

\begin{framed}
\ctext[RGB]{233,212,201}{A} \ctext[RGB]{238,135,106}{triumph} \ctext[RGB]{243,197,175}{,} \ctext[RGB]{239,206,188}{relentless} \ctext[RGB]{246,187,160}{and} \ctext[RGB]{232,119,93}{beautiful} \ctext[RGB]{247,176,146}{in} \ctext[RGB]{243,150,120}{its} \ctext[RGB]{246,164,134}{downbeat} \ctext[RGB]{216,219,225}{darkness} \ctext[RGB]{225,218,214}{.} \vspace{0.1in}

\ctext[RGB]{240,139,109}{Ranks} \ctext[RGB]{226,217,212}{among} \ctext[RGB]{246,188,162}{Willams} \ctext[RGB]{236,210,196}{'} \ctext[RGB]{232,119,93}{best} \ctext[RGB]{247,174,145}{screen} \ctext[RGB]{244,157,126}{work} \ctext[RGB]{242,200,179}{.} \vspace{0.1in}

\ctext[RGB]{245,192,167}{Zany} \ctext[RGB]{231,214,204}{,} \ctext[RGB]{232,119,93}{exuberantly} \ctext[RGB]{179,204,250}{irreverent} \ctext[RGB]{113,148,244}{animated} \ctext[RGB]{208,218,233}{space} \ctext[RGB]{245,160,129}{adventure} \ctext[RGB]{146,180,254}{.}\vspace{0.1in}

\ctext[RGB]{217,220,224}{Behind} \ctext[RGB]{245,191,165}{the} \ctext[RGB]{243,198,176}{snow} \ctext[RGB]{194,212,243}{games} \ctext[RGB]{187,209,247}{and} \ctext[RGB]{237,132,103}{lovable} \ctext[RGB]{246,182,154}{Siberian} \ctext[RGB]{236,210,196}{huskies} \ctext[RGB]{246,164,134}{(} \ctext[RGB]{230,215,207}{plus} \ctext[RGB]{210,218,231}{one} \ctext[RGB]{141,175,253}{sheep} \ctext[RGB]{202,216,238}{dog} \ctext[RGB]{240,204,185}{)} \ctext[RGB]{224,218,215}{,} \ctext[RGB]{178,203,251}{the} \ctext[RGB]{160,191,254}{picture} \ctext[RGB]{241,144,114}{hosts} \ctext[RGB]{168,197,253}{a} \ctext[RGB]{133,168,251}{parka-wrapped} \ctext[RGB]{244,154,123}{dose} \ctext[RGB]{247,174,145}{of} \ctext[RGB]{232,119,93}{heart} \ctext[RGB]{151,184,254}{.} \vspace{0.1in}

\ctext[RGB]{216,219,225}{...} \ctext[RGB]{232,119,93}{a} \ctext[RGB]{239,137,108}{haunting} \ctext[RGB]{244,154,123}{vision} \ctext[RGB]{225,218,214}{,} \ctext[RGB]{246,169,138}{with} \ctext[RGB]{234,211,199}{images} \ctext[RGB]{245,193,168}{that} \ctext[RGB]{193,212,244}{seem} \ctext[RGB]{205,217,236}{more} \ctext[RGB]{247,181,152}{like} \ctext[RGB]{180,205,250}{disturbing} \ctext[RGB]{167,196,253}{hallucinations} \ctext[RGB]{241,203,184}{.} \vspace{0.1in}

\ctext[RGB]{113,148,244}{Suffocated} \ctext[RGB]{242,200,179}{at} \ctext[RGB]{230,215,207}{conception} \ctext[RGB]{210,218,231}{by} \ctext[RGB]{164,194,254}{its} \ctext[RGB]{139,174,253}{Munchausen-by-proxy} \ctext[RGB]{247,178,149}{mum} \ctext[RGB]{189,210,246}{.}\vspace{0.1in}

\ctext[RGB]{179,204,250}{It} \ctext[RGB]{208,218,233}{'s} \ctext[RGB]{245,192,167}{an} \ctext[RGB]{146,180,254}{awfully} \ctext[RGB]{113,148,244}{derivative} \ctext[RGB]{245,160,129}{story} \ctext[RGB]{231,214,204}{.}\vspace{0.1in}

\ctext[RGB]{189,210,246}{A} \ctext[RGB]{113,148,244}{dreadful} \ctext[RGB]{210,218,231}{live-action} \ctext[RGB]{139,174,253}{movie} \ctext[RGB]{164,194,254}{.}

\end{framed}
    \caption{More \estar{} explanations. The shade of background color represents feature importance. }
    \label{fig:qual-examples-app}
\end{figure}

Fig.~\ref{fig:qual-examples-app} presents more visualizations of E$^*$ explanations. These examples suggest that E$^*$ mostly focus on words that convey strong sentiments, which is a plausible working mechanism of a sentiment classifier. 

\section{An Analysis on the \rankins{} Metric}
\label{app:rank-ins}

As introduced in App.~\ref{sec:background}, \rankdel{} evaluates the monotonicity of the model prediction curve when more important features are successively inserted into an empty input. While this expectation seems reasonable, it suffers from a critical issue due to the convention in ranking features: if a feature contributes \textit{against} the prediction, such as a word of sentiment opposite to the prediction (e.g., a positive prediction on ``Other than the story plot being a bit \ul{boring}, everything else is actually masterfully designed and executed.''), it should have negative attribution and the convention is to put them lower in the rank (i.e., less important) than those have zero contributions. This implementation leads to the correct interpretation of all other metric values. 

However, under this convention, the first few words added to the empty input should decrease the model prediction and then increase it, leading to a U-shaped curve. In fact, it is the comprehensiveness curve shown in Fig.~\ref{fig:comp-suff}, flipped both horizontally (because features are inserted rather than removed) and vertically (because the plotted quantity is the model prediction rather than change in prediction). Thus, a deeper U-shape should be preferred, but it is less monotonic. This also explains why the random attribution baseline achieves such a high ranking correlation: as we randomly add features from the empty string to the full input, on average the curve should be a more or less monotonic interpolation between model predictions on empty and full inputs, which has better monotonicity rank correlation than the U-shape. 

It is not clear how to fix the metric. Previous works that proposed \citep{luss2021leveraging} or used \citep{chan2022comparative} this metric often ignored the issue. One work \citep{arya2019one} filtered out all features of negative attribution values and evaluate the rank correlation only on the rest. This, however, is easily manipulatable by an adversary. Specifically, an explainer could shift all attribution values down such that only the most positive one has a non-negative value. This change results in a perfect correlation as long as removing most positive feature induces a decrease in model prediction -- an especially low requirement to satisfy. Empirically, we found that inserting features based on their (unsigned) magnitude barely affects the result either. Thus, we argue that this metric is not a good measurement of explanation quality. 

\section{Visualization of Perturbation Effects}
\label{app:crossing}
\begin{figure}[!htb]
    \centering
    \includegraphics[width=0.8\textwidth]{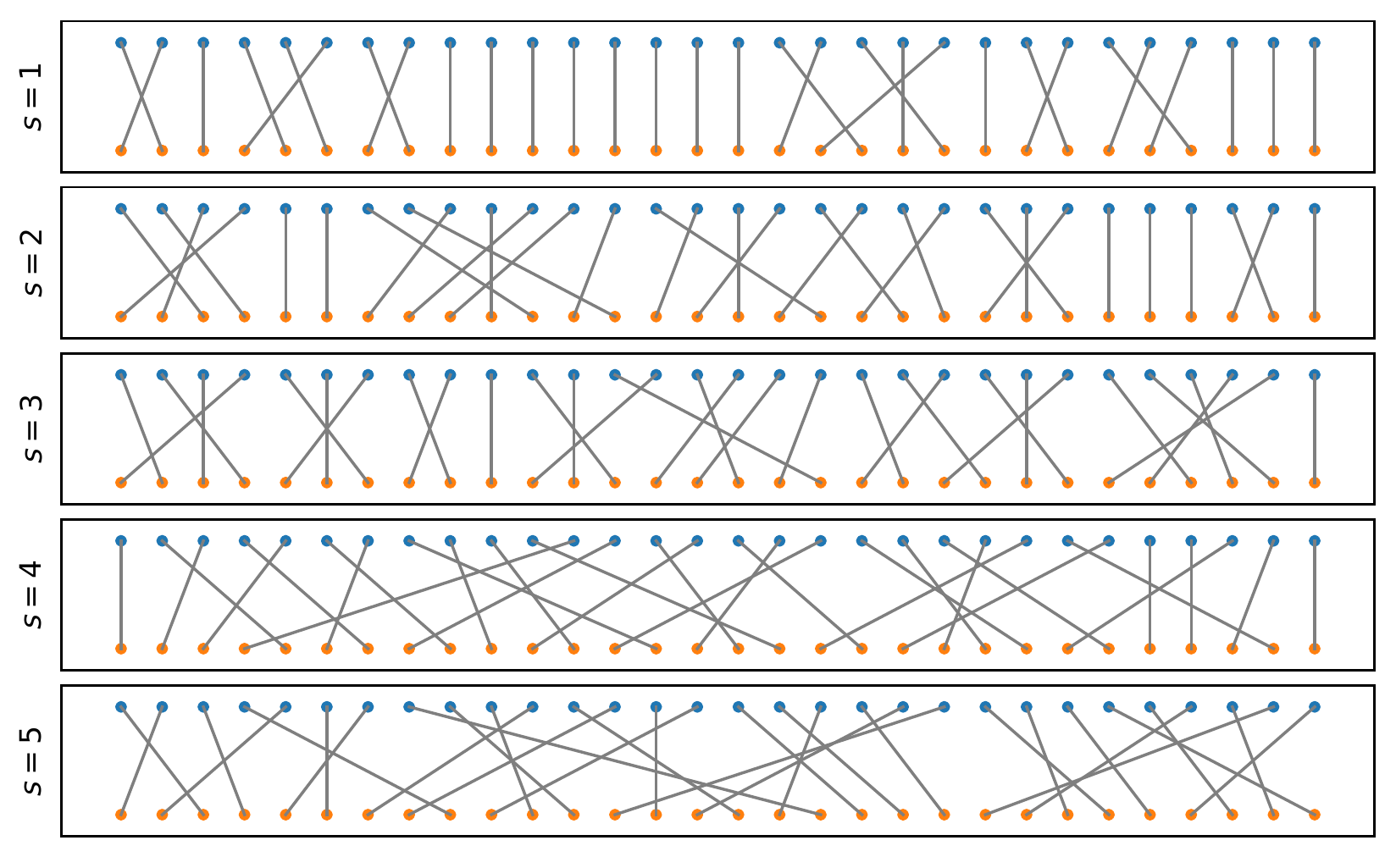}
    \caption{Visualization of rank perturbation under different values of $s$. }
    \label{fig:crossing}
\end{figure}

Fig.~\ref{fig:crossing} visually presents the random perturbation, with different standard deviation $s$ of the Gaussian noise. In each panel, the top row orders the features by their ranked importance, from least important on the left to most on the right, and the bottom row orders the features with perturbed ranked importance, with lines connecting to their original position. For example, in the top panel for $s=1$, the perturbation swaps the relative order of the two least important features on the left. 

\section{Another Assessment on the Explainer-Attacking-Model Behavior}
\label{app:perturbation}

\begin{figure}[!b]
    \centering
    \includegraphics[width=0.55\columnwidth]{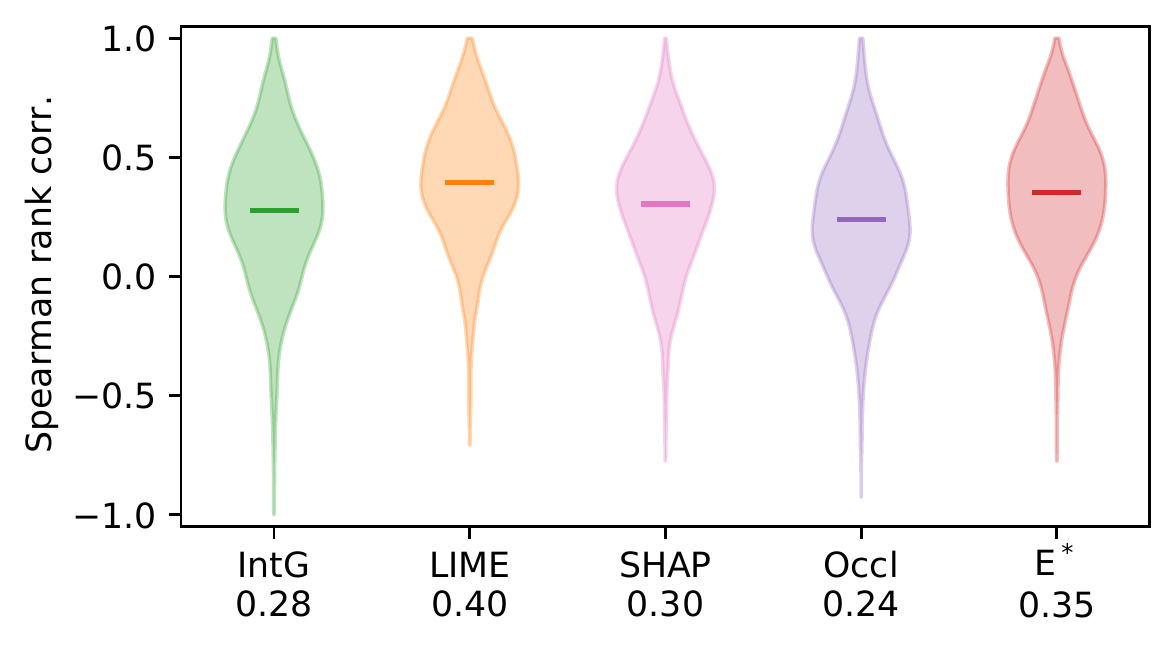}
    \caption{Spearman rank correlation coefficient between intrinsic word polarity score and attribution value. }
    \label{fig:understandability-rank}
\end{figure}

We describe another experiment to assess whether the explanations exploit the adversarial vulnerability of the model. While it is possible that the model could use some shortcuts \citep{geirhos2020shortcut}, we would expect it to predominantly use sentiment-conveying words, as it achieves high accuracy and no such shortcuts are known for the dataset. In this case, we should expect an explainer that does not adversarially exploit the model to give attributions for words correlated with their sentiment values, while an explainer that attacks the model would rate words that are ``adversarial bugs'' to be more important. 

Conveniently, the SST dataset provides human annotations of the polarity score between 0 and 1 for each word, where 0 means very negative, 1 means very positive, and 0.5 means neutral. We compute the alignment between the attribution values (for the positive class) and this score for each word. Given a sentence $x = (x_1, ..., x_L)$ with explanation $e = (e_1, ..., e_L)$ and word polarity score $s = (s_1, ..., s_L)$, the alignment is defined as the Spearman rank correlation coefficient $\rho(e, s)$. Since the vanilla gradient only produces non-negative values, it is impossible to identify whether a word contributes \textit{to} or \textit{against} the positive class, and we exclude it from the analysis. 

Fig.~\ref{fig:understandability-rank} plots the distribution of rank correlations among the test set instances, with the average shown as the bar and also annotated on the horizontal axis. Although no method achieves very high alignment, \estar{} is the second-highest, after LIME. Thus, giving out assumption that high-polarity words are the indeed genuine signals used by the model for making predictions, we can conclude that \estar{} does not adversarially exploit the model for its vulnerability any more severely than the heuristic explainers.

\end{document}